\documentclass{elsarticle}%
\usepackage{amsmath,amssymb,amsfonts}
\DeclareMathOperator*{\argmax}{arg\,max}
\usepackage{algorithmic}
\usepackage{graphicx}
\usepackage{adjustbox}
\usepackage{subfig}
\usepackage{float}
\usepackage{textcomp}
\usepackage{xcolor}
\usepackage{siunitx}
\usepackage{comment}
\usepackage[colorinlistoftodos,prependcaption,textsize=tiny]{todonotes}

\usepackage{lineno,hyperref}

\begin{document}

\title{FuCiTNet: Improving the generalization of deep learning networks by the fusion of learned class-inherent transformations}




\author[1]{Manuel Rey-Area}
\author[2]{Emilio Guirado}
\author[3]{Siham Tabik}
\author[4]{Javier Ruiz-Hidalgo}

\address[1]{atlanTTic Research Center for Telecommunication Technologies, University of Vigo, Galicia, Spain}
\address[2]{Multidisciplinary Institute for Environment Studies "Ramon Margalef”
University of Alicante, Alicante, Spain}
\address[3]{Andalusian Research Institute in Data Science and Computational Intelligence\\ University of Granada 18071,
Spain}
\address[4]{Department of Signal Theory and Communications, Universitat Politècnica de Catalunya, Barcelona, Catalonia, Spain}

\begin{abstract}

It is widely known that very small  datasets produce overfitting in Deep Neural Networks (DNNs), i.e., the network becomes highly biased to the data it has been trained on. This issue is often alleviated using transfer learning, regularization techniques  and/or data augmentation. This work presents a new approach, independent but complementary to the previous mentioned techniques, for improving the generalization of DNNs  on very small datasets in which the involved classes share many visual features. The proposed model, called FuCiTNet (Fusion Class inherent Transformations Network), inspired by GANs, creates as many generators as classes in the problem. Each generator, $k$, learns the transformations that bring the input image into the k-class domain. We introduce a classification loss in the generators to drive the leaning of specific k-class transformations. Our experiments demonstrate that the proposed transformations improve  the generalization of the classification model in three diverse datasets.

\end{abstract}
\maketitle

\section{Introduction}

It is well-known that building robust and efficient supervised deep learning networks requires large amounts of quality data, especially when the involved  classes share many  visual features.  Recent technological advances in camera sensors and the potentially unlimited data provided by internet have helped gathering large volumes of images \cite{imageNet, coco_dataset, cifar}. However, labelling such amounts of data is still manual and costly. In consequence, a large number of problems has still to deal with very small labeled datasets and hence, the resulting classification models are usually unable to  generalise correctly to new unseen examples, this problem is known as overfitting.

In general, the issue of overfitting in supervised networks is addressed either from the model or  data point-of-view. In the former, transfer learning or diverse regularization techniques are employed. In the later, increasing the volume of the training set using data augmentation strategies~\cite{alexnet,intro_dataAug2, intro_dataAug3} are considered. For highly sensitive purposes, data augmentation strategies, if not chosen correctly, can potentially change meaningful information resulting in ill-posed training data. Instead of manually selecting data augmentation techniques, recent works showed that learning these transformations from data can lead to significant improvements in the generalization of the models.

Some approaches train the generator of a GAN (Generative Adversarial Network) to learn the suitable augmenting techniques  from scratch \cite{learn_dataAug, intro_learnAug3} while others train the generator to learn finding the optimal set of augmenting techniques from an initial space of data augmentation strategies \cite{learn_dataAug_gans}. The downside of the aforementioned approaches is that the selected transformations  are applied  to all samples in a training set without taking into account the particularities of each  class.

The present work proposes FuCiTNet approach for  learning class-inherent transformations in each class within the dataset under evaluation. FuCiTNet is inspired by GANs, it creates a number, $N$, of generators equal to the number of classes. Each generator, $k$, with $k \in \{1,\ldots,N\}$,  will be entrusted to learn the features of a specific $k$-class space. When a sample is fed into the system, it is broadcasted to every generator producing $N$ transformed images, each of which is fed to the classifier which predicts a label with certain error. The error is transferred back to the entrusted generator (specified by the input's label ground-truth) indicating the amount of change the class transformation must be altered to meet the classifier requirements. The final prediction of the trained classification model will be calculated based on the  fusion of the N different output predictions.

The contributions of this work can be summarized as follows:
\begin{itemize}
    \item We propose class-inherent transformation generators for improving the generalization capacity of image classification models, especially appropri- ate for problems in which the involved classes share many visual features. Our approach, FuCiTNet, creates as many generators as the number of classes N. Each generator, $k \in \{1,\ldots,N\}$, learns the inherent transformations that bring the input image from its space to the k-class space. We introduce a classification loss in the generators to drive the learning of specific k-class transformations.

    
    \item The final prediction of the  classification model is calculated as a fusion of the N output scores. The source code of FuCiTNet will be available in Github after acceptation
    
    \item Our experiments  demonstrate  that  class-inherent transformations produce a clearer discrimination for the classifier yielding better generalisation performance in three small datasets from two different fields.
    
    \end{itemize}

This paper is organized as follows. A summary of the  most related works to ours are reviewed in Section 2. A description of  FuCiTNet model is provided in Section 3. Experimental framework is provided in Section 4. Results and analysis are given in Section 5 and finally conclusions and future work in Section 6.

\section{Related Work}
Improving the performance of supervised deep neural networks in image classification is still  ongoing research. 
To reach high accuracies the model needs to generalise robustly to unseen cases to eventually avoid overfitting. This is addressed using several approaches.

The most popular approach is data augmentation. It was firstly introduced by Y. LeCun \textit{et al\@.}~\cite{first_data_aug} by randomly applying  these transformations to the training dataset: shearing, random cropping and rotations. With the revolution of CNNs \cite{alexnet}, novel transformations appeared, such as horizontal flipping, changes in intensity, zooming, noise injection etc. With data augmentation, most CNN-based classifiers  reduce overfitting and increase their classification accuracy. 

Dropout, a well known regularization technique, is also  used for improving generalization. It was first introduced by Srivastava \textit{et al\@.}~\cite{dropout}. The key idea is to randomly drop units (along with their connections) from the neural network throughout the training process. This prevents units from co-adapting too much to the data. It significantly reduces overfitting and gives major improvements over other regularization methods.

The emergence of GANs~\cite{gans} has led to promising results in image classification. They were useful for generating synthetic samples to increase small datasets where the number of samples per class was low eventually introducing variability for the generalisation of the classification models. The downside of GANs is that its latent space converges if there exists a fair good amount of images to train. Generating synthetic samples in small and very small dataset is still an open issue. 

For small and very small datasets, the problem of overfitting is even greater. Inspired by GANs, the authors in~\cite{learn_dataAug} designed an  augmentation network that learns the set of augmentations that best improve the classifier performance. The classifier tells the generator network which configuration of image transformations  prefers when distinguishing samples from different classes. In the same direction, the authors in~\cite{learn_dataAug_rl,learn_dataAug_gans} address a similar problem using Reinforcement Learning and adversarial strategy respectively. The former chooses from a list of potential transformations which one is the best suited through augmentation policies. The latter combined the list of transformations to synthesize a total image transformation. The present work is different to all the previously cited  works in that it proposes a new approach for learning transformations inherent to each particular class based on the classifier requirements eventually forcing the classes to be as distinguishable as possible from each other.


\section{FuCiTNet approach}

 Inspired by GANs, FuCiTNet learns class-inherent transformations.  Our aim is to build a generator  that improves the discrimination capacity between the different classes. 
GANs use the so called adversarial loss which optimizes a min-max problem. The generator, $G$, tries to minimize the following function while the discriminator, $D$, tries to maximize it:
\begin{equation}
\begin{split}
\min_{G}\max_{D} \ \mathbb{E}_{X\sim p_{data(X)}}\left[\log D(X)\right]+ \\ 
\mathbb{E}_{z\sim p_z{(z)}}\left[\log(1-D(G(z))\right]
\end{split}
\end{equation}


Where  $D(X)$ is the discriminator's estimate of the probability that real data instance $X$ is real. $\mathbb{E}_{X\sim p_{data(X)}}$ is the expected value over all real data instances. $G(z)$ is the generator's output  given a noise $z$. $D(G(z))$ is the discriminator's estimate of the probability that a fake instance is real. $\mathbb{E}_{z\sim p_z{(z)}}$ is the expected value over all random inputs to the generator. The formula derives from the cross-entropy between the real and generated distributions.

In this work, the adversarial concept is slightly changed. Instead of using a discriminator that focuses on maximizing the inter-class probability of the image belonging to the data or latent distribution, we use a classifier that minimizes the loss of $G(\mathrm{X})$ of belonging to a particular dataset class. As the images produced by the generator do not need to be similar to its ground truth, we restrict the generator latent space in a different way  by replacing $D(X)$ with the classification loss of $G(X)$, which will hallucinate samples with specific enhanced features to meet the classifier requirements improving the classification accuracy. 

\begin{figure*}[h]
\centerline{\includegraphics[trim={2cm 13.5cm 2cm 2cm},clip, width=0.85\textwidth, height=0.45\textheight] {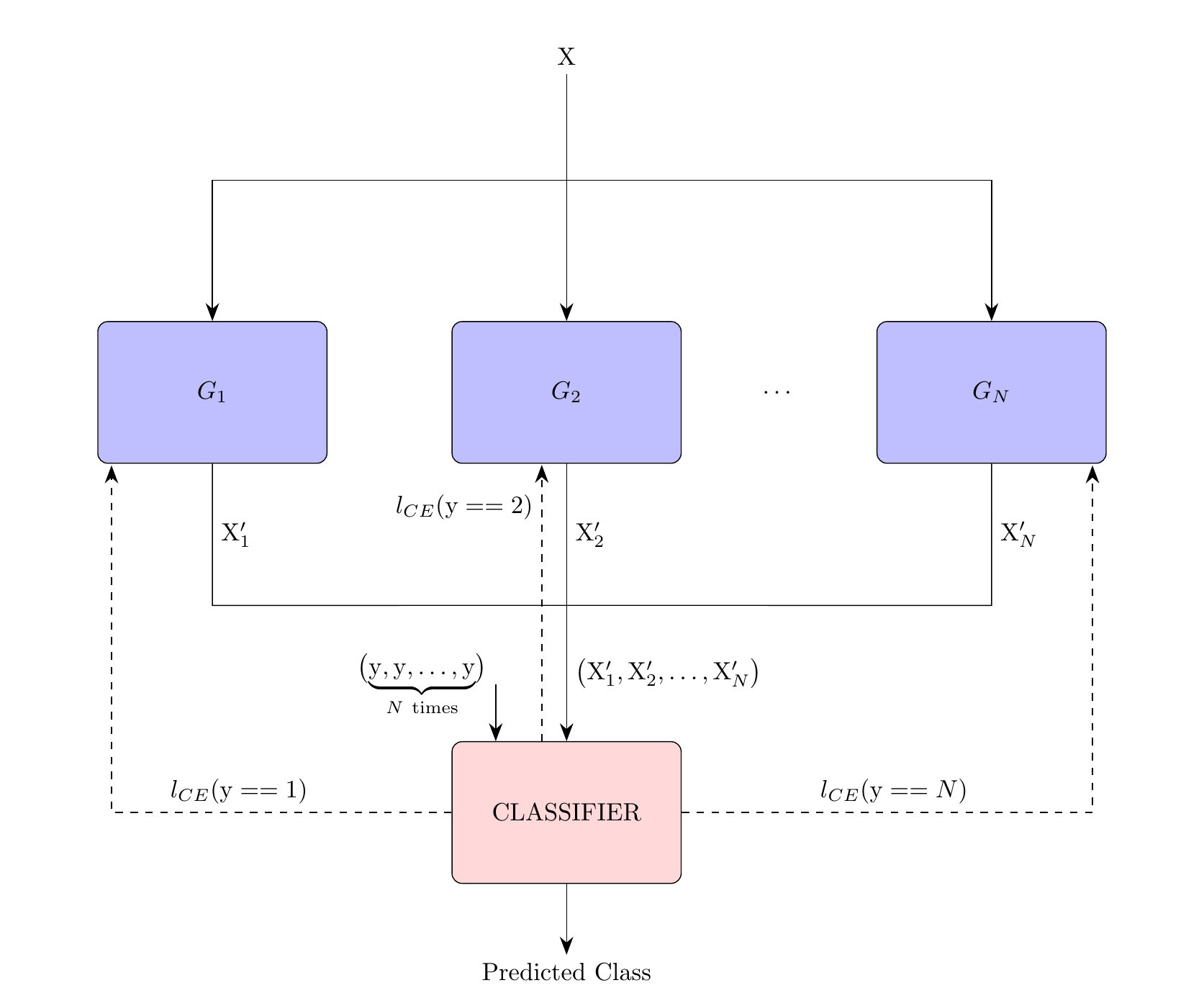}}
\caption{Flowchart of FuCiTNet during training. The input image $\mathrm{X}$ is broadcasted to every generator. Each generator produces a transformed image $\mathrm{X}^{\prime}$. The classifier computes the cross entropy loss which is transferred back to the generator commissioned to enhance features of the class given by the input's groundtruth label $\mathrm{y}$.}
\label{fig:workflow_train}
\end{figure*}

Instead of building one single generator $G$, we create an array of $N$ generators, $G_k$ with $k \in \{1,\ldots,N\}$. Where $N$ is equal the number of object classes. Each generator  $G_k(\mathrm{X})=\mathrm{X^{\prime}_k}$ is in charge of learning  the inherent features of that specific k-class until $\mathrm{X'}$ becomes part of the k-class space. In other words, each generator $G_k$ will learn the transformations that map the input image $\mathrm{X}$ from its own i-class domain to the k-class domain, with $i \in \{1,\ldots,N\}$. The flowchart diagram of FuCiTNet is depicted in Figure \ref{fig:workflow_train}.

\begin{figure*}[h]
\centerline{\includegraphics[trim={2cm 13.5cm 2cm 2cm},clip, width=0.85\textwidth, height=0.45\textheight] {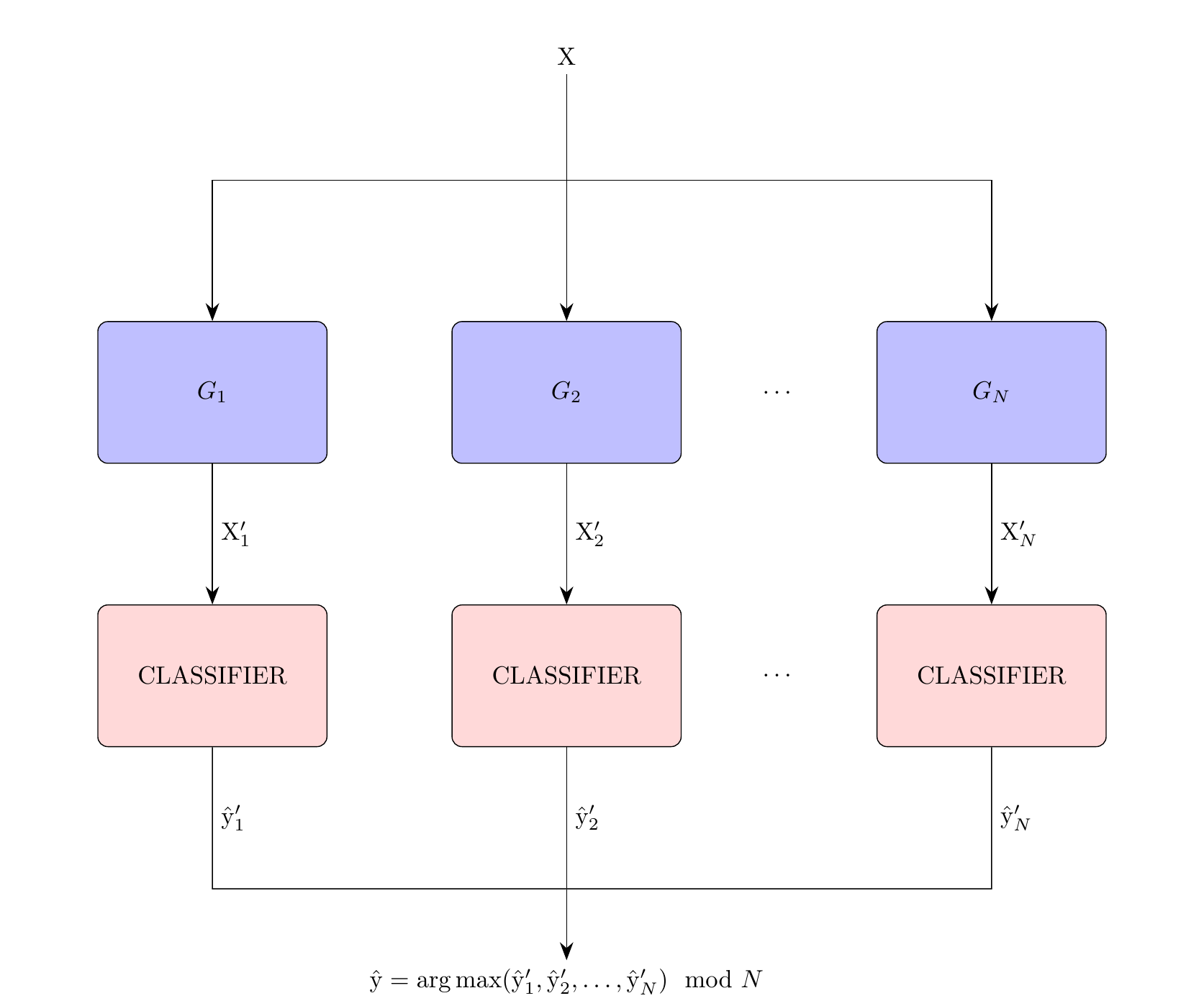}}
\caption{Flowchart of FuCiTNet in inference.}
\label{fig:workflow_test}
\end{figure*}

The architecture of our generators $G_k$, with $k \in \{1,\ldots,N\}$,  consists of 5 identical residual blocks \cite{residual_nets}.  Each block has two convolutional layers with $3\times3$ kernels and 64 feature maps followed by batch-normalization layers \cite{batch_norm} and ParametricReLU \cite{param_relu} as  activation function. The last residual block is followed by a final convolutional layer which reduces the output image channels to 3 to match the input's dimensions.

Our classifier is a ResNet-18 which consists of an initial convolutional layer with $7 \times 7$ kernels and 64 feature maps followed by a $3 \times 3$ max pool layer. Then, 4 blocks of two convolutional layers with $3\times3$ kernels with 64, 128, 256 and 512 feature maps respectively followed by a $7\times 7$ average pooling and a fully connected layer which outputs a $N$ element vector. ReLU is used as the activation function \cite{relu}. 

The classifier uses a standard cross entropy loss ($l_{CE}$). Given a batch size $B$ of input images with their respective ground truth  labels,
$\{\mathrm{X}^{(j)},\mathrm{y}^{(j)} | \  j=1,\ldots,B\}$:
 
\begin{equation}
    l_{CE} = -\frac{1}{B}\sum_{j=1}^{B}\sum_{k=1}^{N}\sum_{i=1}^{N}{\mathrm{y}^{(j)}[i] \cdot \log(\mathrm{\hat{y}}_{k}^{\prime(j)}[i])}
\label{eq:cross_entropy}
\end{equation}
The true label distribution for the $j^{th}$ image is depicted by $\mathrm{y}^{(j)}$ and $\mathrm{\hat{y}}^{\prime (j)}_k$ indicates the predicted label distribution for the transformed image $\mathrm{X}^{\prime(j)}_k$.

All generators use the same loss inspired by \cite{SRGAN}. For the $k^{th}$ generator with $k \in \{1,\ldots,N\}$ the respective loss $\mathcal{L}_{gen_k}$ is indicated by Formula (\ref{eq:system_loss}). It consists of a multiple term loss constituted by a pixel-wise MSE term, a perception MSE term and the classifier loss. Adding this classifier loss to the generators is one of the main novelty of our method. The classification loss is added to the generator loss with a weighted factor $\lambda$ indicating how much the generator must change its outcome to suit the classifier. It is worth noting that, for each input image, the classification loss is only transferred to a particular generator indicated by the input's ground-truth label.

\begin{equation}
   \mathcal{L}_{gen_k} = \underbrace{l_{MSE} + 0.006\cdot l_{Perceptual}}_\text{Similarity term: ${\mathcal{L}_{sim}}$} + \lambda\cdot l_{CE}(\mathrm{y}==k)
\label{eq:system_loss}
\end{equation}

The similarity term (${\mathcal{L}_{sim}}$) keeps the generator from changing the image too much while the classifier loss drives the output away from the input and close to the k-class feature space.

The pixel-wise MSE is obtained by performing a regular L2-norm between each pixel in the input image and the generated image. 
\begin{equation}
    l_{MSE} = \frac{1}{WH}\sum_{x=1}^W\sum_{y=1}^H(\mathrm{X}_{x,y} - \mathrm{X}_{x,y}^{\prime})^2
\end{equation}

The perceptual MSE assesses similarity between $\mathrm{X}$ and $\mathrm{X}'$ by feeding each of them into a pretrained VGG-16 \cite{vgg}. The euclidean distance between the $\mathrm{j}^{th}$ VGG-16 feature maps ($\phi_n$) defines the perceptual loss. 
\begin{equation}
    l_{perceptual} = \frac{1}{W_jH_j}\sum_{x=1}^{W_j}\sum_{y=1}^{H_j}(\phi_j(\mathrm{X}_{x,y})-\phi_j(\mathrm{X}_{x,y}^{\prime}))^2
\end{equation}

The cross entropy loss coming from the classifier is added to a particular generator loss in the generator array specified by the input's ground-truth label $\mathrm{y}$ with a weighted factor $\lambda$ controlling the impact of the classifier within the generator:
\begin{equation}
l_{CE}(\mathrm{y}==k) =  -\frac{1}{B}\sum_{j=1}^{B}\sum_{k=1}^{N}{\mathrm{y}^{(j)}[k] \cdot \log(\mathrm{\hat{y}}_{k}^{\prime(j)}[k])}
\end{equation}
where $B$ is the batch size, $N$ the number of classes and $\mathrm{y}^{(j)}[k]$ is the element in the true label distribution of the $j^{th}$ image belonging to class $k$. Likewise, $\mathrm{\hat{y}}_{k}^{\prime(j)}[k]$ depicts the classifier predictions for the transformed image $\mathrm{X}^{\prime(j)}_k$ with $k \in \{1,\ldots,N\}$ of belonging to the class $k$.

In this manner, each  generator in the array is able to learn class-inherent features from a specific class eventually, accentuating the differentiation among classes.

In inference time the general flowchart of the system is modified as shown in Figure~\ref{fig:workflow_test}. The final class prediction is given by concatenating each output distribution $\mathrm{\hat{y}}^{\prime}$ for every $\mathrm{X^{\prime}}$ in the logit domain, taking the arg max and computing the modulo with the amount of classes $N$ as indicated in Formula (\ref{eq:final_pred})

\begin{equation}
    \mathrm{\hat{y}} = \argmax(\mathrm{\hat{y}}^{\prime}_1,\mathrm{\hat{y}}^{\prime}_2,\ldots ,\mathrm{\hat{y}}^{\prime}_N)\mod N
\label{eq:final_pred}
\end{equation}

\section{Experimental framework}

To assess the capacity of FuCiTNet in increasing the discrimination among classes,  we created three  datasets from two different fields using object-classes that are frequently confused by the best performing classification model. In particular, from Tiny ImageNet \cite{le2015tiny}, we created two datasets made of two and three classes respectively,  cat-vs-dog and cat-vs-dog-vs-goldfish. From  NWPU-RESISC45  remote sensing dataset made of aerial ortho-images \cite{Cheng_2017}, we created church-vs-palace dataset which represent one of the most similar/confusing pair of classes in this dataset.  To further increase the complexity of this dataset, we downsampled the church and palace images to $64 \times 64$ pixels. A brief description of the three datasets is provided in Table \ref{tab:data_dist}.
 
\begin{table}[h]
\centering
\resizebox{\textwidth}{!}{%
\begin{tabular}{|l||c|c|c|c|c|}
\hline
{\textbf{Dataset}} & \textbf{\# classes} & \textbf{\# pixels/image} &\textbf{Train} & \textbf{Test} & \textbf{Total} \\ \hline\hline
cat-vs-dog                           & two  &$64 \times 64$ & 800            & 200           & 1000           \\\hline
cat-vs-dog-vs-goldfish              & three & $64 \times 64$& 1200           & 300           & 1500           \\\hline
church-vs-palace                     & two  & $64 \times 64$& 1120           & 280           & 1400  \\\hline        
\end{tabular}}
\caption{Description of the three evaluated datasets together with the used data distribution 80\%-20\%.}
\label{tab:data_dist}
\end{table}

The $N$ generator networks were initialized randomly while the classifier, ResNet-18, was initialized using the pretrained weights on ImageNet \cite{imageNet}. As optimization algorithm, we used Adam \cite{adam} in both, generator and classifier with $\beta_1=0.9$. The adopted learning rates for generators and classifier are different, for the generators, we used a value of $10^{-4}$ and $10^{-3}$ for  ResNet-18. The reason why generators have a lower learning rate is because they need to be  subtle when generating the transformed image and avoid the classifier from falling behind on capturing the rate of change in appearance. For the ResNet-18 we used learning rate decay of 0.1 each 5 epochs. We have also used weight decay of $10^{-4}$ to avoid overfitting. We applied early stopping monitoring based on the validation loss with a patience of 10 epochs.

The weighted factor $\lambda$ in the generator loss in Eq. (\ref{eq:system_loss}) is a hyperparameter to be tuned. We evaluated a space of 13 values: [1, 0.5, 0.1, 0.075, 0.05, 0.025, 0.01, 0.0075, 0.005, 0.0025, 0.001, 0.00075, 0.0005]. We assessed the effect from a high contribution in the loss towards a softer impact. For each value of $\lambda$ the system was trained throughout 100 epochs with a batch size of 32. We alternate updates between the generator and classifier, i.e.,  for each batch we update first the classifier then we update the generator.
We evaluate FuCiTNet on the validation set after each epoch. The chosen weights for both networks are the ones which minimize the classification loss in this set.



For a fair comparison, we compare the results of FuCiTNet {\color{black} with the two most related approaches  ~\cite{learn_dataAug} and \cite{learn_dataAug_gans}. As \cite{learn_dataAug_gans} provides the source code, we analyze and show the results on the three considered datasets, cat-vs-dog, cat-vs-dog-vs-goldfish and church-vs-palace, following the same experimental protocol we used in the rest of experiments. However, as  \cite{learn_dataAug} does not provide the source code and considered only binary classes, we included only the results  reported in the paper using the same experimental protocol on  dog-vs-cat dataset. The results of this approach on cat-vs-dog-vs-goldfish and church-vs-palace are not available to us. Both approaches from \cite{learn_dataAug} and \cite{learn_dataAug_gans} used ResNet CNN architecture. In addition, we also compare our results to the results of the best classification model obtained based on the same network architecture and by manually selecting the set of optimizations that reaches the highest performance. }

In all the experiment, 
we  used 3-fold cross validation  following a 80:20 hold out data distribution as depicted in Table \ref{tab:data_dist}. All the  experiments were executed on a NVIDIA Titan Xp. All the implementations were performed using PyTorch   DL framework \cite{pytorch}. 

\section{Results and analysis}
This section presents, compares and analyzes the quantitative and qualitative results of FuCiTNet with the state-of-the-art methods, \cite{learn_dataAug} and \cite{learn_dataAug_gans},
 and  with the best classification model.

\begin{table}[H]
\centering
\resizebox{\textwidth}{!}{%
\begin{tabular}{|l|c|c|c|}
\hline
\multicolumn{1}{|c|}{Setup}                                             & \textbf{Accuracy} & \multicolumn{1}{l|}{ Cat mean} & \multicolumn{1}{l|}{ Dog mean} \\ 
\multicolumn{1}{|c|}{ }                                                 & \textbf{ } & \multicolumn{1}{l|}{ confidence} & \multicolumn{1}{l|}{ confidence} \\ 
\hline\hline
None                                                                        & 0.872                              & ---                                      & ---                                      \\ \hline
FT                                                                          & 0.865                              & ---                                      & ---                                      \\ \hline
Data aug                                                                    & 0.892                              & 1.673                                    & 2.520                                    \\ \hline
Data aug, FT                                                                & 0.888                              & ---                                      & ---                                      \\ \hline
\cite{learn_dataAug}  & 0.770                              & ---                                      & ---   \\  \hline     
\cite{learn_dataAug_gans}  & 0.800                              & ---                                      & ---   \\  \hline     
{FuCiTNet, None, $\lambda = 0.01$}         & 0.870                              & ---                                      & ---                                      \\ \hline
{FuCiTNet,  FT, $\lambda = 0.025$}         & 0.880                              & ---                                      & ---                                      \\ \hline
{FuCiTNet, Data aug,  $\lambda = 0.0075$}  & 0.883                              & ---                                      & ---                                      \\ \hline
{FuCiTNet, Data aug, FT, $\lambda = 0.05$} & \textbf{0.912}    & 2.734                                    & 2.101                                    \\ \hline
\end{tabular}}
\caption{\color{black} Performance of ResNet-18 classification model without (row: 1 to 4) and with (row: 7 to 10) FuCiTNet, using different configurations,  on cat-vs-dog dataset. Data aug consists of random horizontal flipping, random rotation and random affine. The accuracy of the state-of-the-art approaches, \cite{learn_dataAug} and \cite{learn_dataAug_gans}, is shown in row 5 and 6 respectively.}
\label{tab:cat-dog}
\end{table}

\begin{table}[h]
\subfloat[Confusion matrix for the setup: Data aug]{
\resizebox{0.476\textwidth}{!}{%
\begin{tabular}{|l|c|c|}\hline
Predicted/Actual & Cat & Dog \\ \hline
Cat              & 86  & 14  \\
Dog              & 11  & 89 \\\hline
\end{tabular}}}
\quad
\subfloat[Confusion matrix for the setup: FuCiTNet, Data aug, FT, $\lambda=0.05$]{
\resizebox{0.476\textwidth}{!}{%
\begin{tabular}{|l|c|c|}\hline
Predicted/Actual & Cat & Dog \\ \hline
Cat              & 97  & 3   \\
Dog              & 10  & 90  \\\hline
\end{tabular}}}
\caption{Cat-vs-dog confusion matrices for fold \#1 in 3FCV for the best reference model(a) and FuCiTNet (b)}
\label{confusion:cat-dog}
\end{table}

\begin{table}[H]
\centering
\resizebox{\textwidth}{!}{%
\begin{tabular}{|l|c|c|c|}
\hline
\multicolumn{1}{|c|}{Setup}                                             & \textbf{Accuracy} & \multicolumn{1}{l|}{ \footnotesize  Church mean} & \multicolumn{1}{l|}{  Palace mean} \\ 
\multicolumn{1}{|c|}{ }                                                 & \textbf{ } & \multicolumn{1}{l|}{  confidence} & \multicolumn{1}{l|}{ confidence} \\ 
\hline\hline
None                                                                          & 0.774                              & ---                                         & ---                                         \\ \hline
No data aug, FT                                                               & 0.769                              & ---                                         & ---                                         \\ \hline
Data aug                                                                      & 0.783                              & 0.769                                       & 1.452                                       \\ \hline
Data aug, FT                                                                  & 0.779                              & ---                                         & ---                                         \\ \hline
\cite{learn_dataAug_gans}  & 0.779                              & ---                                      & ---   \\  \hline  
{FuCiTNet, None, $\lambda = 0.05$}           & 0.751                              & ---                                         & ---                                         \\ \hline
{FuCiTNet,  FT, $\lambda = 0.0025$}          & 0.767                              & ---                                         & ---                                         \\ \hline
{FuCiTNet, Data aug,  $\lambda = 0.01$}      & 0.777                              & ---                                         & ---                                         \\ \hline
{FuCiTNet, Data aug, FT, $\lambda = 0.0025$} & \textbf{0.795}    & 1.868                                       & 1.580                                       \\ \hline
\end{tabular}}
\caption{\color{black} Performance of ResNet-18 based classification model without (row: 1 to 4) and with (row: 6 to 9) FuCiTNet, using different configurations, on the church-vs-palace dataset. Data augmentation consists of random horizontal flipping, random rotation and random affine.  The accuracy of the state-of-the-art approach  \cite{learn_dataAug_gans} is shown in row  5.}
\label{tab:chursh-vs-palace}
\end{table}

\begin{table}[h]
\subfloat[Confusion matrix for the setup: Data aug]{
\begin{adjustbox}{max width=0.475\textwidth}
\begin{tabular}{|l|c|c|}\hline
Predicted/Actual & Church & Palace \\ \hline
Church           & 114    & 26     \\ 
Palace           & 30     & 110   \\\hline
\end{tabular}
\end{adjustbox}}
\quad
\subfloat[Confusion matrix for the setup: FuCiTNet, Data aug, FT, $\lambda=0.0025$]{
\begin{adjustbox}{max width=0.475\textwidth}
\begin{tabular}{|l|c|c|}\hline
Predicted/Actual & Church & Palace \\ \hline
Church           & 114    & 26     \\
Palace           & 27     & 113   \\\hline
\end{tabular}
\end{adjustbox}}
\caption{Church-vs-palace confusion matrices for fold \#1 in 3FCV for the best reference model(a) and FuCiTNet (b)}
\label{confusion:church-palace}
\end{table}

\begin{table}[H]
\centering
\resizebox{\textwidth}{!}{%
\begin{tabular}{|l|c|c|c|c|}
\hline
\multicolumn{1}{|c|}{Setup}                                                  & \textbf{Accuracy} & {  Cat mean} & { Dog mean} & {Goldfish mean} \\ 
          & & { confidence} & {  confidence} & { confidence}\\\hline\hline

None                                                                         & 0.898                              & ---      & ---      & ---           \\ \hline
FT                                                                           & 0.909                              & ---      & ---      & ---           \\ \hline
Data aug                                                                     & 0.915                              & 7.796    & 4.778    & 9.617         \\ \hline
Data aug, FT                                                                 & 0.911                              & ---      & ---      & ---           \\ \hline
\cite{learn_dataAug_gans}  &  0.790    & ---      & ---  & --- \\  \hline  
{FuCiTNet, None, $\lambda = 0.0075$}        & 0.887                              & ---      & ---      & ---           \\ \hline
{FuCiTNet,   FT, $\lambda = 0.0075$}        & 0.902                              & ---      & ---      & ---           \\ \hline
{FuCiTNet, Data aug,  $\lambda = 0.005$}    & 0.900                              & ---      & ---      & ---           \\ \hline
{FuCiTNet, Data aug, FT, $\lambda = 0.005$} & \textbf{0.920}    & 3.484    & 3.248    & 6.613         \\ \hline
\end{tabular}}
\caption{\color{black}  Performance of ResNet-18 based classification model without (row: 1 to 4) and with (row: 6 to 9) FuCiTNet, using different configurations, on cat-vs-dog-vs-goldfish dataset. Data augmentation consists of random horizontal flipping, random rotation and random affine. The accuracy of the state-of-the-art approach  \cite{learn_dataAug_gans} is shown in row 5.}
\label{tab:cat-vs-dog-vs-goldfish}
\end{table}

\begin{table}[h]
\subfloat[Confusion matrix for the setup: Data aug]{
\begin{adjustbox}{max width=0.475\textwidth}
\begin{tabular}{|l|c|c|c|}\hline
Predicted/Actual & Cat & Dog & Goldfish \\ \hline
Cat              & 94  & 6   & 0        \\
Dog              & 18  & 80 & 2        \\
Goldfish         & 3   & 2   & 95     \\\hline
\end{tabular}
\end{adjustbox}}
\quad
\subfloat[Confusion matrix for the setup: FuCiTNet, Data aug, FT, $\lambda=0.005$]{
\begin{adjustbox}{max width=0.475\textwidth}
\begin{tabular}{|l|c|c|c|}\hline
Predicted/Actual & Cat & Dog & Goldfish \\ \hline
Cat              & 94  & 6   & 0        \\
Dog              & 9   & 85  & 6        \\
Goldfish         & 1   & 1   & 98  \\\hline
\end{tabular}
\end{adjustbox}}
\caption{Cat-vs-dog-vs-goldfish confusion matrices for fold \#2 in 3FCV for the best reference model(a) and FuCiTNet(b)}
\label{confusion:cat-dog-goldfish}

\end{table}


\paragraph{Quantitative results} The  performance, in terms of accuracy,   of the best classification model with and without applying FuCiTNet,  on the three datasets is presented in Tables \ref{tab:cat-dog}, \ref{tab:chursh-vs-palace} and \ref{tab:cat-vs-dog-vs-goldfish}. Several configurations were analyzed, 'None' indicates that the network was initialized with ImageNet weights and retrained  on the dataset. 'Data aug' indicates that the set of the optimal data augmentation techniques was applied. 'FT', for Fine Tuning, indicates that only the last fully connected layer was retrained on the dataset and the remaining layers were frozen. If 'FT' is not indicated, it means that the whole layers of the network were re-trained on the dataset. {\color{black}To compare FuCiTNet with the state-of-the-art, we also include the accuracy of  the method proposed in   

\cite{learn_dataAug} on cat-vs-dog dataset only (see Table \ref{tab:cat-dog}) and the method proposed in \cite{learn_dataAug_gans} on the three datasets (see Tables \ref{tab:cat-dog}, \ref{tab:chursh-vs-palace} and \ref{tab:cat-vs-dog-vs-goldfish}
).}

When applying FuCiTNet to the test images,  the classification model reaches
$2.24\%$, $1.53\%$ and $0.54\%$ higher accuracy  than the best classification model on cat-dog, church-vs-palace and cat-vs-dog-vs-goldfish respectively. The number of TP and TN also  improves in all the three datasets as shown in the confusions matrices in Tables \ref{confusion:cat-dog},  \ref{confusion:church-palace} and \ref{confusion:cat-dog-goldfish}. 

{\color{black}With respect to the state-of-the-art, FuCiTNet provides 14\%, 2\% and 16.4\% better accuracy than  the approach proposed in 
 \cite{learn_dataAug_gans} on  cat-dog, church-vs-palace and cat-vs-dog-vs-goldfish respectively and  18.44\% better accuracy 
 than the approach proposed by \cite{learn_dataAug}  on cat-vs-dog.}

These results can  explain that FuCiTNet makes the object-class in the input images  more distinguishable to the model. Our transformation approach can be considered as an incremental optimization to the rest of optimizations space, since  on all the  datasets, the best results were obtained by combining FuCiTNet  with fine-tuning and data-augmentation.

We have also analyzed the mean confidence of each class in the three problems such that:

$$\text{mean confidence}(k\text{-class})=\frac{\sum_{i=1}^{T} P_k(i)}{T}$$

where $T$ is the  number of correctly classified test images, from the test split, with $k$ ground truth class. The mean confidence of the best reference model and FuCiTNet are shown in the third column of   Tables \ref{tab:cat-dog}, \ref{tab:chursh-vs-palace} and \ref{tab:cat-vs-dog-vs-goldfish}. As we can observe from these results FuCiTNet clearly improve the mean confidence of the model in all the classes in cat-vs-dog and church-vs-palace datasets. Although FuCiTNet have lowered the mean confidence  in the cat-vs-dog-vs-goldfish dataset,  the accuracy, number of true positives and true negatives have improved.

FuCiTNet provides several advantages over the method proposed in \cite{learn_dataAug} and \cite{learn_dataAug_gans}. It does not require neither paired datasets of high and low resolution input images nor super-resolution network. Unlike \cite{learn_dataAug}, in which the transformations were learnt regardless the class of the sample, our results demonstrate that exploiting class-inherent features  improves significantly the borders  between visually similar classes.




\begin{figure}[H]
\centering
\subfloat[$(2.862, -3.777)$]{\includegraphics[trim={0 0 0 0},clip, width=0.27\textwidth, height=0.2\textheight] {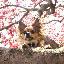}}
\hspace{0.05cm}
\subfloat[$(2.393, -2.376)$]{\includegraphics[trim={0 0 0 0},clip, width=0.27\textwidth, height=0.2\textheight] {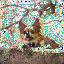}}
\hspace{0.05cm}
\subfloat[$(2.239, -2.212)$]{\includegraphics[trim={0 0 0 0},clip, width=0.27\textwidth, height=0.2\textheight] {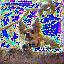}}
\hspace{0.05cm}
\subfloat[$(1.816, -2.261)$]{\includegraphics[trim={0 0 0 0},clip, width=0.27\textwidth, height=0.2\textheight] {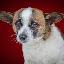}}
\hspace{0.05cm}
\subfloat[$(2.912, -2.960)$]{\includegraphics[trim={0 0 0 0},clip, width=0.27\textwidth, height=0.2\textheight] {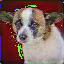}}
\hspace{0.05cm}
\subfloat[$(2.037, -2.007)$]{\includegraphics[trim={0 0 0 0},clip, width=0.27\textwidth, height=0.2\textheight] {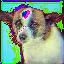}}
\caption{Left) Original sample of dog class; middle) Dog transformation to original sample; right) Cat transformation to original sample. The confidence score of the model for (dog class, cat class) are indicated below each image. The score for the left image is obtained using the best reference model whereas the ones for the middle and right images are obtained with FuCiTNet.}
\label{fig:dog}
\end{figure}

\begin{figure}[H]
\centering
\subfloat[$(-0.631, 0.664)$]{\includegraphics[trim={0 0 0 0},clip, width=0.27\textwidth, height=0.2\textheight] {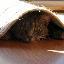}}
\hspace{0.05cm}
\subfloat[$(2.561, -2.557)$]{\includegraphics[trim={0 0 0 0},clip, width=0.27\textwidth, height=0.2\textheight] {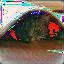}}
\hspace{0.05cm}
\subfloat[$(3.597, -3.679)$]{\includegraphics[trim={0 0 0 0},clip, width=0.27\textwidth, height=0.2\textheight] {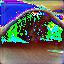}}
\hspace{0.05cm}
\subfloat[$(-8.048, 6.966)$]{\includegraphics[trim={0 0 0 0},clip, width=0.27\textwidth, height=0.2\textheight] {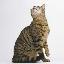}}
\hspace{0.05cm}
\subfloat[$(-1.625, 1.961)$]{\includegraphics[trim={0 0 0 0},clip, width=0.27\textwidth, height=0.2\textheight] {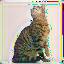}}
\hspace{0.05cm}
\subfloat[$(-0.202, 0.433)$]{\includegraphics[trim={0 0 0 0},clip, width=0.27\textwidth, height=0.2\textheight] {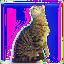}}
\caption{left) Original sample of cat class; middle) Dog transformation to original sample; right) Cat transformation to original sample. The confidence score of the model for (dog class, cat class) are indicated below each image. The score for the left image is obtained using the best reference model whereas the ones for the middle and right images are obtained with FuCiTNet.}
\label{fig:cat}
\end{figure}

\begin{figure}[H]
\centering
\subfloat[$(0.936, 0.104)$]{\includegraphics[trim={0 0 0 0},clip, width=0.27\textwidth, height=0.2\textheight] {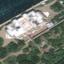}}
\hspace{0.05cm}
\subfloat[$(0.099, -0.362)$]{\includegraphics[trim={0 0 0 0},clip, width=0.27\textwidth, height=0.2\textheight] {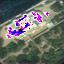}}
\hspace{0.05cm}
\subfloat[$(-0.512, 0.278)$]{\includegraphics[trim={0 0 0 0},clip, width=0.27\textwidth, height=0.2\textheight] {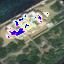}}
\hspace{0.05cm}
\subfloat[$(-0.279, 0.530)$]{\includegraphics[trim={0 0 0 0},clip, width=0.27\textwidth, height=0.2\textheight] {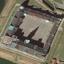}}
\hspace{0.05cm}
\subfloat[$(-0.556, 0.550)$]{\includegraphics[trim={0 0 0 0},clip, width=0.27\textwidth, height=0.2\textheight] {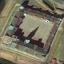}}
\hspace{0.05cm}
\subfloat[$(-0.677, 0.661)$]{\includegraphics[trim={0 0 0 0},clip, width=0.27\textwidth, height=0.2\textheight] {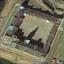}}
\caption{left) Original sample of Palace class; middle) Church transformation to original sample; right) Palace transformation to original sample. The confidence score of the model for (church class, palace class) are indicated below each image. The score for the left image is obtained using the best reference model whereas the ones for the middle and right images are obtained with FuCiTNet.}
\label{fig:palace}
\end{figure}

\begin{figure}[H]
\centering
\subfloat[$(-0.356,0.644)$]{\includegraphics[trim={0 0 0 0},clip, width=0.27\textwidth, height=0.2\textheight] {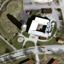}}
\hspace{0.05cm}
\subfloat[$(0.534,-0.674)$]{\includegraphics[trim={0 0 0 0},clip, width=0.27\textwidth, height=0.2\textheight] {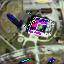}}
\hspace{0.05cm}
\subfloat[$(-0.295,0.167)$]{\includegraphics[trim={0 0 0 0},clip, width=0.27\textwidth, height=0.2\textheight] {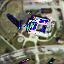}}
\hspace{0.05cm}
\subfloat[$(0.422,0.646)$]{\includegraphics[trim={0 0 0 0},clip, width=0.27\textwidth, height=0.2\textheight] {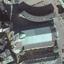}}
\hspace{0.05cm}
\subfloat[$(0.626,-0.670)$]{\includegraphics[trim={0 0 0 0},clip, width=0.27\textwidth, height=0.2\textheight] {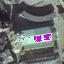}}
\hspace{0.05cm}
\subfloat[$(0.249,-0.291)$]{\includegraphics[trim={0 0 0 0},clip, width=0.27\textwidth, height=0.2\textheight] {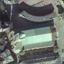}}
\caption{left) Original sample of Church class; middle) Church transformation to original sample; right) Palace transformation to original sample. The confidence score of the model for (church class, palace class) are indicated below each image. The score for the left image is obtained using the best reference model whereas the ones for the middle and right images are obtained with FuCiTNet.}
\label{fig:church}
\end{figure}

\begin{figure}[H]
\centering
\subfloat[$(\text{-}0.91,\text{-}2.78,3.68)$]{\includegraphics[trim={0 0 0 0},clip, width=0.25\textwidth, height=0.16\textheight]
{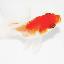}}
\subfloat[$(\text{-}6.06,\text{-}1.10,5,57)$]{\includegraphics[trim={0 0 0 0},clip, width=0.25\textwidth, height=0.16\textheight] {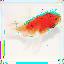}}
\subfloat[$(\text{-}6.71,\text{-}1.17,5.99)$]{\includegraphics[trim={0 0 0 0},clip, width=0.25\textwidth, height=0.16\textheight] {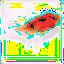}}
\subfloat[$(\text{-}5.41,\text{-}1.11,5.03)$]{\includegraphics[trim={0 0 0 0},clip, width=0.25\textwidth, height=0.16\textheight] {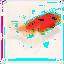}}
\hspace{0.0001cm}
\subfloat[$(\text{-}1.41,\text{-}1.56,2.82)$]{\includegraphics[trim={0 0 0 0},clip, width=0.25\textwidth, height=0.16\textheight] {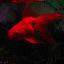}}
\subfloat[$(\text{-}3.05,\text{-}0.83,3.77)$]{\includegraphics[trim={0 0 0 0},clip, width=0.25\textwidth, height=0.16\textheight] {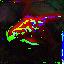}}
\subfloat[$(\text{-}2.99,\text{-}0.80,3.68)$]{\includegraphics[trim={0 0 0 0},clip, width=0.25\textwidth, height=0.16\textheight] {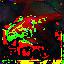}}
\subfloat[$(\text{-}3.12,\text{-}0.88,3.91)$]{\includegraphics[trim={0 0 0 0},clip, width=0.25\textwidth, height=0.16\textheight] {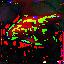}}
\caption{left) Original sample of goldfish class; middle left) Cat transformation to original sample; middle right) Dog transformation to original sample; right) Goldfish transformation to original sample. The confidence score of the model for (dog class, cat class, goldfish class) are indicated below each image. The score for the left image is obtained using the best reference model whereas the others are obtained with FuCiTNet.}
\label{fig:cat_vs_dog_vs_goldfish}
\end{figure}

\paragraph{Qualitative results} Figures \ref{fig:dog}, \ref{fig:cat}, \ref{fig:palace}, \ref{fig:church} and \ref{fig:cat_vs_dog_vs_goldfish} show visually the transformations applied by FuCiTNet to different original input images from different classes. As it can be observed from these images, the transformations affect: 

\begin{itemize}
    \item the contour or border of the object class in the transformed images, as it can be seen in Figures \ref{fig:dog}(c) and (f) and Figures \ref{fig:cat}(c) and (f).
    
    \item the pixels that constitute the body of the object-class, as it is the case  in Figure \ref{fig:dog}(f) and Figures \ref{fig:palace}(b) and (c), Figures \ref{fig:church}(b), (c) and (e), Figures \ref{fig:cat_vs_dog_vs_goldfish}(c), (d), (f), (g) and (h).
    
    \item or background as it is the case in Figures  \ref{fig:dog}(c) and (f), Figures \ref{fig:cat}(b), (c), (e) and (f).
\end{itemize}

In some images in the church-vs-palace testset, the model does not add any transformation. This  indicates that the classifier does not need any additional transformation to differentiate the object-class in those specific images. In general, in this dataset, most of the palaces have either a dome or a rectangular roof with an inner courtyard but generally the churches do not include any of these features. In the cases where a church has an inner courtyard, the model needs to add a transformation so that the classifier can  distinguish it better from a church as it can be seen in Figure \ref{fig:church}. Likewise, when the palace has a very distinguishable shape from a church, the model does not add any transformation as it can be seen in Figure \ref{fig:palace}(d), (e) and (f).

This effect does not occur in the cat-vs-dog and cat-vs-dog-vs-fish test sets, the background in these images is so diverse, it includes, people, furniture, occlusion etc.,  that the model always need some more transformation to differentiate better between the involved classes.

\section{Conclusions and future work}
Our aim in this work was  reducing over-fitting  in very small datasets in which the involved  object-classes share many visual features. We presented FuCiTNet model in which a novel array of generators  learn independently class-inherent transformations  of each class of the problem. We introduced a classification loss in the generators to drive the learning of specific k-class transformations. The learnt transformations are then applied to the input test images to help the classifier distinguishing better between different classes.  

Our experiments demonstrated that FuCiTNet increases the classification  generalization capability on three small datasets with very similar classes. With the benchmark datasets we demonstrate that FuCiTNet behaves robustly in diverse-nature data and handles properly different view dimensions (zenital and frontal). We conclude that our method yields strong gains as an incremental optimization technique additional to the standards when searching for a better model performance.

As future work, we are planning to explore and adapt FuCiTNet to small medical datasets.

\bibliographystyle{elsarticle-num}
\bibliography{mybib}
\end{document}